\pdfoutput=1

\documentclass[11pt]{article}

\usepackage[final]{acl}

\usepackage{times}
\usepackage{latexsym}
\usepackage{tabularx}
\usepackage{booktabs}
\usepackage{multirow}
\usepackage{listings}
\usepackage{float}
\usepackage{caption}
\usepackage{makecell}
\usepackage{tabularx}
\usepackage{amsmath}

\usepackage[T1]{fontenc}

\usepackage[utf8]{inputenc}

\usepackage{microtype}

\usepackage{inconsolata}

\usepackage{graphicx}
\definecolor{ao}{rgb}{0.0, 0.5, 0.0}

\title{The Role of Exploration Modules in Small Language Models for Knowledge Graph Question Answering}

\author{Yi-Jie Cheng$^{1,2}\quad$Oscar Chew$^1\quad$Yun-Nung Chen$^2$ \\
  $^1$ASUS \\
  $^2$National Taiwan University \\
  \texttt{b09202004@ntu.edu.tw$\quad$oscar\_chew@asus.com$\quad$y.v.chen@ieee.org}
}

\begin{document}
\maketitle
\begin{abstract}
Integrating knowledge graphs (KGs) into the reasoning processes of large language models (LLMs) has emerged as a promising approach to mitigate hallucination. However, existing work in this area often relies on proprietary or extremely large models, limiting accessibility and scalability. In this study, we investigate the capabilities of existing integration methods for small language models (SLMs) in KG-based question answering and observe that their performance is often constrained by their limited ability to traverse and reason over knowledge graphs. To address this limitation, we propose leveraging simple and efficient exploration modules to handle knowledge graph traversal in place of the language model itself. Experiment results demonstrate that these lightweight modules effectively improve the performance of small language models on knowledge graph question answering tasks. Source code: \url{https://github.com/yijie-cheng/SLM-ToG/}.
\end{abstract}

\section{Introduction}
Large Language Models such as GPT4 \cite{openai2024gpt4}, Gemini \cite{google2023gemini}, Qwen \cite{qwen} have achieved state-of-the-art performance across a wide range of natural language processing tasks. Despite their impressive capabilities, a key limitation is the lack of interpretability in their decision-making processes. Moreover, they are prone to hallucination, especially when the required knowledge is not present in their parametric memory. To tackle these challenges, Think-on-Graph (ToG; \citealp{sun2024thinkongraph}) treats the LLM as an agent that dynamically interacts with knowledge graphs to retrieve external knowledge, exemplifying a LLM$\times$KG paradigm that has garnered significant attention. To cast LLMs as an agent,  ToG and similar approaches typically rely on very large models \cite{xu-etal-2024-generate, cheng-etal-2024-call, liang2025fast}, limiting their accessibility for low-resource settings. Other recent efforts \cite{luo2024reasoning,he2024g,ao2025lightprof,yang-etal-2025-curiousllm} have proposed additional reasoning or exploration modules to improve LLM-KG integration, but these methods require task-specific training or fine-tuning.

In this paper, we focus on a practical setting where end users or system deployers have access only to small- or medium-sized language models for inference. In this context, an important question arises: how effectively can these SLMs leverage knowledge graphs for question answering? To explore this, we examine Think-on-Graph \cite{sun2024thinkongraph}, a representative training-free framework, and observe that when applied to SLMs rather than LLMs, ToG underperforms and sometimes even falls behind the Chain-of-Thought (CoT) baseline \cite{wei2022chain}. Through detailed analysis, we attribute this failure to the SLMs' limited ability to explore and reason over knowledge graphs. 
We argue that using lightweight passage retrieval methods such as SentenceBERT and GTR for exploration can substantially enhance the effectiveness of knowledge graph traversal for SLMs. We would like to point out that the novelty of our work does not lie in introducing new models or architectures. Rather, we revisit previously underestimated techniques and demonstrate their effectiveness in enhancing reasoning performance in resource-constrained settings. Our contributions can be summarized as follows:
\begin{itemize}
    \item We demonstrate that the existing ToG framework is not as effective for SLMs in KGQA.
    \item We identify the exploration stage as a key bottleneck for SLM performance in knowledge graph reasoning.
    \item We show that incorporating simple and efficient passage retrieval modules significantly improves SLMs' ability to traverse and reason over knowledge graphs.

\end{itemize}

\section{Traversing Knowledge Graphs with Small Language Models}
\subsection{Preliminaries}

Think-on-Graph \cite{sun2024thinkongraph} is a framework for KGQA that casts a language model as an agent navigating a knowledge graph to perform multi-hop reasoning. It operates in three main stages:
\begin{itemize}
    \item Initialization: The model extracts topic entities from the input question and locates them in the KG to form initial reasoning paths.
    \item Exploration: Using beam search, the model iteratively expands these paths by exploring neighboring relations and entities. At each step, the LLM ranks candidates and prunes less relevant options, guided by the question context.
    \item Reasoning: Once sufficient evidence is gathered, the LLM generates a final answer based on the maintained reasoning paths.
\end{itemize}

This structured interaction enables interpretable and context-sensitive reasoning while leveraging the strengths of both KGs and language models.

\subsection{Exploration Modules for SLMs}
In Section~\ref{sec:bottleneck}, we will show that SLMs are less effective for KGQA due to their limitation in exploration stage. To address the weaknesses of using only SLM itself for exploration of KG, we examine the use of simple, efficient retrieval models in Section~\ref{sec:slm_retrieval}. These models, which measure semantic similarity between text segments, have shown strong performance in passage retrieval tasks and hence are well-suited to assist SLMs in pruning irrelevant candidates during KG traversal. Importantly, they can be used in a zero-shot, plug-and-play manner, requiring no additional training or fine-tuning, making them well-suited for low-resource settings.
\paragraph{Classic Retrieval Index} BM25 \cite{robertson2009bm25} is a ranking function used in information retrieval that scores how well a document matches a query based on term frequency and how common the term is across all documents. \paragraph{Dense Retrieval} We consider two dense retrievers: SentenceBERT \cite{reimers-gurevych-2019-sentence}, a BERT-based model fine-tuned for producing semantically meaningful sentence embeddings, and GTR \cite{ni-etal-2022-large}, a T5-based model optimized for passage retrieval tasks. Both models have approximately 110 million parameters which is substantially smaller than the smallest SLM (0.5B) evaluated in this work. Implementation details are presented in Appendix.~\ref{sec:implementation}.

\section{Experiments}

\begin{table}
  \centering
  \begin{tabular}{llcc}
    \toprule
    \bf Models & & \bf CWQ & \bf WebQSP \\
    \midrule
    \multicolumn{3}{l}{\it Large Language Models} \\
    GPT-4.1 & w/ CoT & 0.457 & 0.710 \\
     & w/ ToG & \textbf{0.540} & \textbf{0.813} \\ 
\midrule
    \multicolumn{3}{l}{\it Small Language Models} \\
    Qwen2-0.5b & w/ CoT & \textbf{0.129} & \textbf{0.272} \\
    & w/ ToG & 0.081 & 0.210 \\ 
    Gemma2-2b & w/ CoT & 0.127 & 0.373 \\
    & w/ ToG & \textbf{0.140} & \textbf{0.382} \\ 
    Phi-3-mini-3.8b& w/ CoT & \textbf{0.273} & \textbf{0.522} \\
    & w/ ToG & 0.270 & 0.520 \\ 
    Qwen2-7b & w/ CoT & 0.275 & 0.544 \\
    & w/ ToG & \textbf{0.300} & \textbf{0.637} \\ 
    Llama-3-8b& w/ CoT & 0.291 & \textbf{0.603} \\
    & w/ ToG & \textbf{0.296} & 0.569 \\ 
    Mean SLM & w/ CoT & \textbf{0.219} & 0.456\\
     & w/ ToG & 0.217 & \textbf{0.464} \\ 
\bottomrule
    
  \end{tabular}
  \caption{Comparison of ToG and CoT across model sizes. While ToG substantially improves GPT-4.1, its effectiveness does not consistently extend to SLMs.}
  \label{tab:cotvstog}
\end{table}

\begin{table*}[ht]
    \centering
  \begin{tabular}{|l|}
\hline
\textbf{Question:} What type of government is used in the country with Northern District? \\ \hline \hline

\textbf{With knowledge triplets retrieved by SLM} \\ \hline

\begin{minipage}[c]{0.1\textwidth}
    \centering
    \includegraphics[scale=0.25]{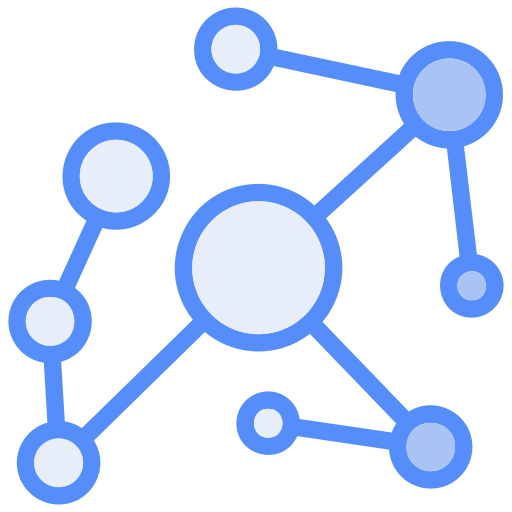}
\end{minipage}%
\hspace{1em}%
\begin{minipage}[c]{0.85\textwidth}
    \makecell[l]{(`Northern District', `country', `Israel'), \\ (`Northern District', `administrative\_parent', `Israel')}
\end{minipage} \\ \hline

\begin{minipage}[c]{0.1\textwidth}
    \centering
    \includegraphics[scale=0.09]{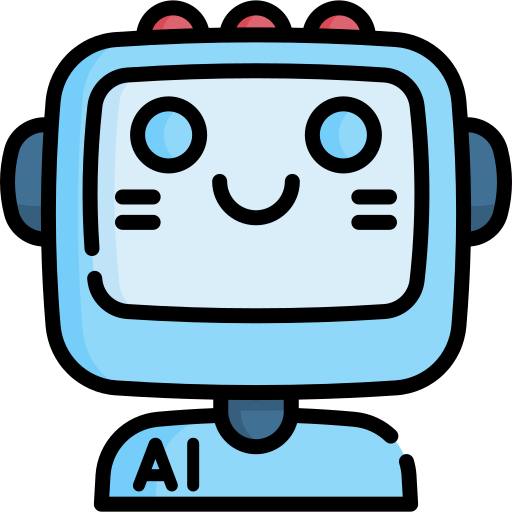}
\end{minipage}%
\hspace{1em}%
\begin{minipage}[c]{0.85\textwidth}
    \makecell[l]{\textbf{SLM:} The triplets do not provide information about the type of government used \\ in Israel.}
\end{minipage} \\ \hline \hline

\textbf{With knowledge triplets retrieved by GPT4.1} \\ \hline

\begin{minipage}[c]{0.1\textwidth}
    \centering
    \includegraphics[scale=0.25]{knowledge-graph.png}
\end{minipage}%
\hspace{1em}%
\begin{minipage}[c]{0.85\textwidth}
    \makecell[l]{ (`Northern District', `country', `Israel'), \\ (`Northern District', `administrative\_parent', 'Israel'), \\ 
\textcolor{ao}{(`Israel', `form\_of\_government', `Parliamentary system')}, \\
(`Israel', `administrative\_children', `Northern District')}
\end{minipage} \\ \hline

\begin{minipage}[c]{0.1\textwidth}
    \centering
    \includegraphics[scale=0.09]{ai.png}
\end{minipage}%
\hspace{1em}%
\begin{minipage}[c]{0.85\textwidth}
    \makecell[l]{\textbf{SLM:} Based on the given knowledge triplets, the country with the Northern District \\ is Israel, which uses a \textcolor{ao}{Parliamentary system} as its form of government.}
\end{minipage} \\ \hline
\end{tabular}
\caption{An example illustrating the limitations of an SLM when performing KG exploration on its own. When relying solely on its retrieved triplets, the SLM fails to answer the question. However, when provided with triplets retrieved by GPT-4.1, including the key relation, the same SLM is able to produce the correct answer.}\label{tab:case}
\end{table*}

\begin{table}[ht]
  \centering
  \begin{tabular}{lcc}
    \toprule
    \bf Models & \bf CWQ & \bf WebQSP \\
    \midrule
    Qwen2-0.5b CoT & 0.129 & 0.272 \\
    $\quad$w/ GPT-4.1 ToG & \textbf{0.301} & \textbf{0.578} \\ \hline
    Gemma2-2b CoT & 0.127 & 0.373\\ 
    $\quad$w/ GPT-4.1 ToG & \textbf{0.306} & \textbf{0.672} \\ \hline
    Phi-3-mini-3.8b CoT & 0.273 & 0.522  \\ 
    $\quad$w/ GPT-4.1 ToG & \textbf{0.421} & \textbf{0.736} \\ \hline
    Qwen2-7b CoT & 0.275 & 0.544 \\ 
    $\quad$w/ GPT-4.1 ToG & \textbf{0.409} & \textbf{0.746} \\ \hline
    Llama-3-8b CoT & 0.291 & 0.603 \\
    $\quad$w/ GPT-4.1 ToG & \textbf{0.451} & \textbf{0.772} \\ \hline
    Improvement w/ GPT4.1 & 0.159 & 0.238 \\
    \bottomrule
  \end{tabular}
  \caption{Performance of SLMs with GPT-4.1-assisted exploration. With high-quality context, SLMs can offer better improvement over the CoT baseline, highlighting exploration as the key bottleneck in the ToG framework}
  \label{tab:cheat}
\end{table}

In this section, we aim to answer the following research questions:
\begin{itemize}
    \item \textbf{RQ1}: How do SLMs perform in KGQA compared to a larger proprietary LLM (GPT-4.1)?
    \item \textbf{RQ2}: Why are SLMs less effective at leveraging knowledge graphs for question answering tasks?
    \item \textbf{RQ3}: How effective are SLMs when paired with better-suited exploration modules?
\end{itemize}
\subsection{Setup}

\paragraph{Datasets and Metrics}
Following \citet{sun2024thinkongraph}, we use Freebase \cite{bollacker2008freebase} as our underlying knowledge graph. We evaluate our models on two benchmark datasets: ComplexWebQuestions (CWQ; \citealp{talmor-berant-2018-web}) and WebQSP \cite{yih-etal-2016-value}. CWQ contains complex questions that require up to 4-hop reasoning while WebQSP which primarily involves 1- to 2-hop reasoning tasks. 
We adopt exact match (EM) score as the primary evaluation metric, which measures whether the predicted answer string exactly matches the given answer.

\paragraph{Language Models}
We consider SLMs ranging in size from 0.5B to 8B parameters. The models include Qwen2 0.5B \cite{yang2024qwen2}, Gemma2-2b \cite{gemma2024gemma}, Phi-3-Mini-3.8B \cite{abdin2024phi}, Qwen2 7b and LLaMA 3-8B \cite{grattafiori2024llama}.




\subsection{RQ1: Think-on-Graph with LLMs and SLMs}
We begin by examining the effectiveness of applying ToG to SLMs in comparison to LLMs. As shown in Table~\ref{tab:cotvstog}, while a giant LLM (GPT-4.1) \footnote{We use the GPT-4.1 snapshot released on April 14, 2025.} enjoys significant boost from ToG, we observe that SLMs equipped with ToG receive limited improvement and can perform even worse than the CoT baseline. This discrepancy underscores a key limitation: while ToG is effective for LLMs, its effectiveness does not translate well to the lower-capacity SLMs with weaker reasoning capabilities.

\subsection{RQ2: Bottleneck of Exploration} \label{sec:bottleneck}

Given that ToG fails to improve performance for SLMs, we further investigate the underlying cause. Our hypothesis is that, without effective exploration, SLMs lack access to the necessary information required to generate correct answers, resulting in low EM scores. To verify this, we test an upper bound where we temporarily assume the access to GPT-4.1 for exploration only. That is, GPT-4.1 is used to explore the knowledge graph and provide context to the SLMs to reason the final outputs. We first look into failure cases of SLMs and found that SLMs could not generate the correct answer due to lack of proper context, as illustrated in Table~\ref{tab:case} \footnote{The figure contains resources from \url{Flaticon.com}}. As shown in Table~\ref{tab:cheat}, with the context provided by GPT-4.1, SLMs are able to reason effectively and offer better improvement over the original CoT baseline. 

\begin{figure}[]
  \centering
  \includegraphics[width=0.99\linewidth]{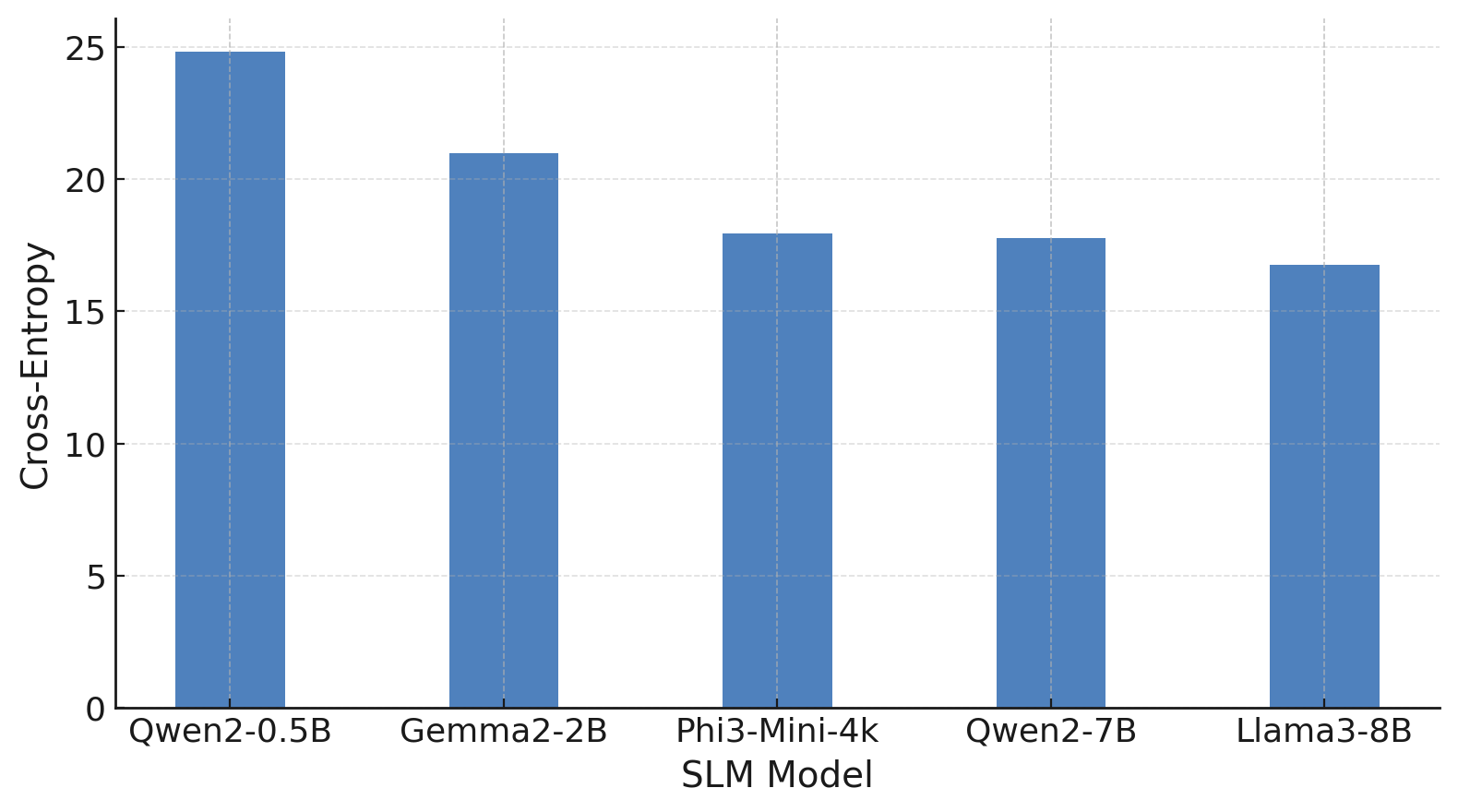}
  \caption{Cross-entropy alignment between the exploration outputs of SLMs and GPT-4.1 across different model sizes. A lower cross-entropy value indicates a closer alignment with GPT-4.1's exploration decisions. The consistent improvement with increasing model size highlights the critical role of exploration quality as the performance bottleneck for SLMs in the ToG framework.}
  \label{fig:exploration_CE}
\end{figure}

We further treat the exploration outputs of GPT-4.1 as pseudo-ground truth and measure how closely the outputs of SLMs align with them in terms of cross-entropy. As shown in Figure~\ref{fig:exploration_CE}, this alignment increases consistently with model size, supporting the view that exploration quality is a key bottleneck for SLMs within the ToG framework.

One might ask whether the difference in performance between SLMs and LLMs are due to their abilities in adhering to the questions/answer format. We have ruled out this possibility by leveraging Constrained Decoding. Relevant details are presented in Appendix~\ref{sec:json}.
\subsection{RQ3: Passage Retrieval for Exploration} \label{sec:slm_retrieval}
\begin{table}[ht]
  \centering
  \begin{tabular}{lcc}
    \toprule
    \bf Models & \bf CWQ & \bf WebQSP \\
    \hline
    Qwen2-0.5b ToG  & 0.081 & 0.210 \\
    $\quad$w/ BM25 & 0.106 & 0.236 \\
    $\quad$w/ SentenceBERT & 0.061 & 0.174 \\
    $\quad$w/ GTR & \textbf{0.123} & \textbf{0.262} \\ \hline
    Gemma2-2b ToG & 0.140 & 0.382 \\
    $\quad$w/ BM25 & 0.127 & 0.236\\
    $\quad$w/ SentenceBERT & 0.136 & \textbf{0.520} \\
    $\quad$w/ GTR & \textbf{0.152} & 0.511 \\ \hline
    Phi-3-mini-3.8b ToG & 0.270 & 0.520 \\
    $\quad$w/ BM25 & 0.254 & 0.416 \\
    $\quad$w/ SentenceBERT & 0.284 & 0.577 \\
    $\quad$w/ GTR & \textbf{0.291} & \textbf{0.605}\\ \hline
    Qwen2-7b ToG   & 0.300 & 0.637 \\
    $\quad$w/ BM25 & 0.251 & 0.513 \\
    $\quad$w/ SentenceBERT & 0.318 & 0.665 \\
    $\quad$w/ GTR & \textbf{0.331} & \textbf{0.671} \\ \hline
    Llama-3-8b ToG & 0.296 & 0.569 \\
    $\quad$w/ BM25 & 0.274 & 0.456 \\
    $\quad$w/ SentenceBERT & 0.313 & 0.640 \\
    $\quad$w/ GTR & \textbf{0.325} & \textbf{0.642} \\ \bottomrule
  \end{tabular}
  \caption{Effectiveness of lightweight passage retrieval methods for KG Exploration. SentenceBERT and GTR provides strong performance gains across models, validating its effectiveness for SLM-based KGQA.}
  \label{tab:retrieval}
\end{table}

As we have determined in Section~\ref{sec:bottleneck} the core limitation of SLMs in the ToG framework lies in their inadequate performance during the exploration stage. One promising direction to address this is to decouple the exploration process from the language model itself. Instead of relying on the SLM to retrieve relevant knowledge paths, we explore the use of lightweight passage retrieval models to assist in this stage. These models are efficient, require no additional training, and have shown strong performance in passage retrieval tasks, making them a natural fit for supporting KG exploration. We present our main results in Table~\ref{tab:retrieval}. Across all SLMs we studied, SentenceBERT and GTR obtain substantial improvement over both the original ToG and CoT for SLMs. This result highlights the effectiveness of leveraging passage retrieval models to assist SLMs during exploration. Interestingly, our findings contrast with those of \citet{sun2024thinkongraph}, who report that integrating passage retrieval models leads to significant performance degradation when applied to LLMs instead of SLMs. We further discuss this in Appendix~\ref{sec:llm_retrieval}.
\section{Conclusion}
In this paper, we investigate the limitations of SLMs in leveraging knowledge graphs for question answering. We identify the core issue as the inadequacy of SLMs in the exploration stage, where they often fail to retrieve accurate reasoning paths and relevant knowledge. To address this, we propose replacing the exploration component in ToG with lightweight passage retrieval models. Experiment results demonstrate that this approach not only improves the efficiency of the reasoning process but also enables SLMs to benefit more effectively from KGs. These findings may serve as a foundation for future research on more effective and accessible use of KGs in practical, real-world settings.

\section*{Limitations}
Due to computational constraints, we report results based on a single run without multiple random seeds. While this prevents us from quantifying variance across runs, the consistent performance trends observed across different models still provide strong evidence supporting our findings.

\section*{Acknowledgments}
We thank the reviewers and the ASUS AIoT team for their valuable feedback.

\bibliography{main}

\appendix
\clearpage

\begin{figure}[ht]
\centering
\includegraphics[width=0.95\linewidth]{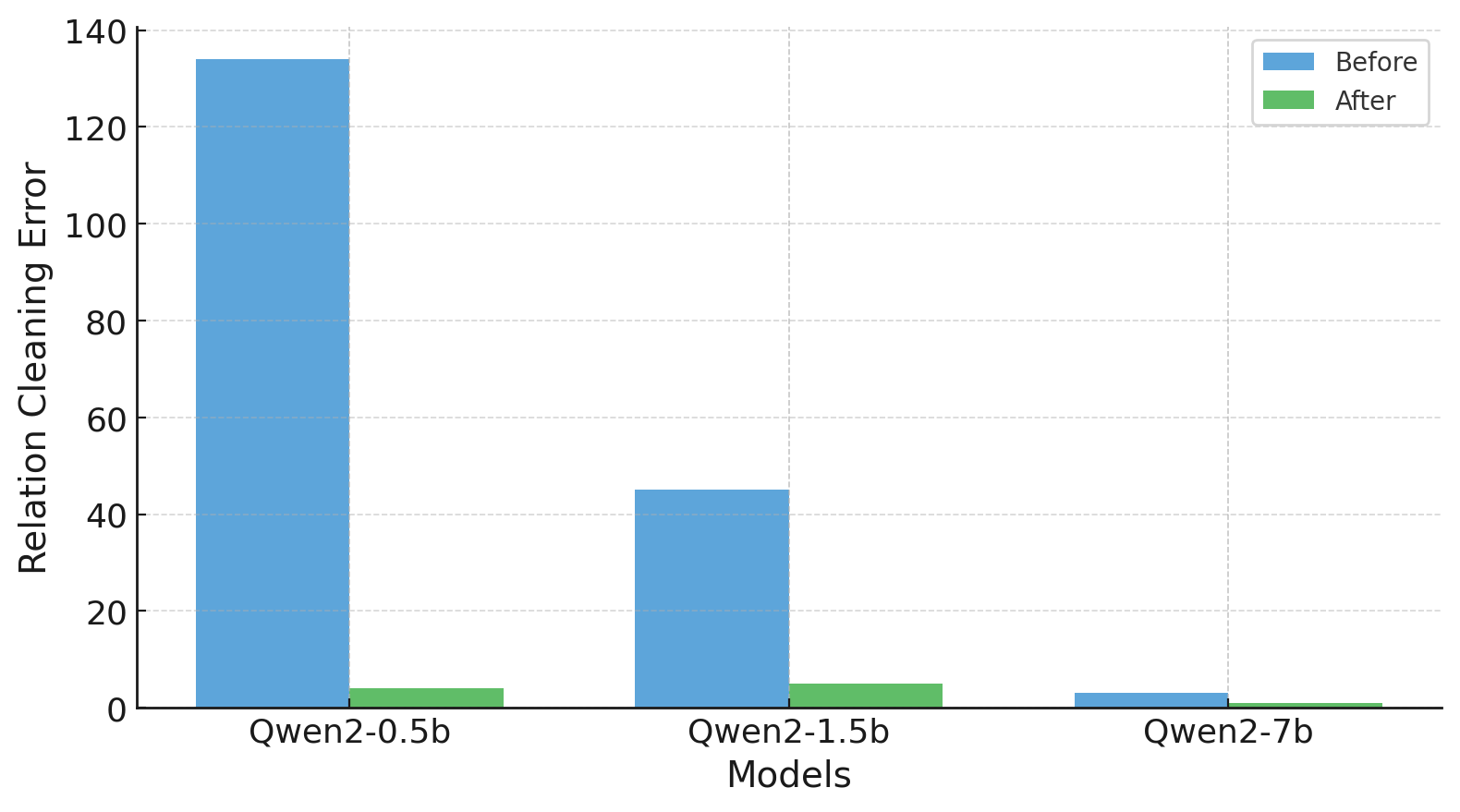}
\caption{Relation cleaning errors before and after applying constrained decoding. Smaller models like Qwen2-0.5b and Qwen2-1.5b show substantial reductions in formatting errors, indicating the effectiveness of our constrained decoding strategy.}
\label{fig:relation_cleaning_error_comparison}
\end{figure}

\begin{figure}[ht]
\centering
\includegraphics[width=0.95\linewidth]{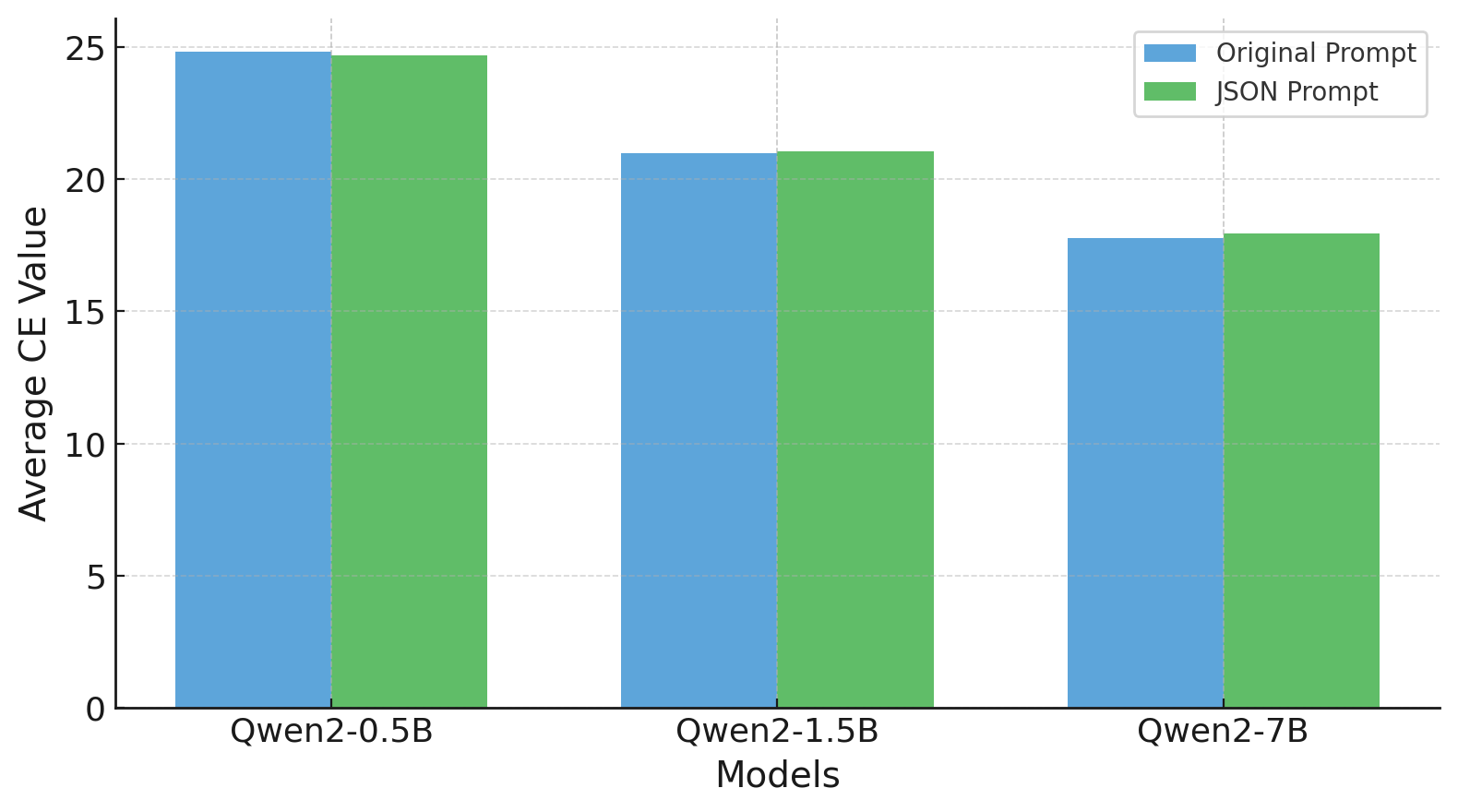}
\caption{Average cross-entropy between model-retrieved relation paths and the pseudo-ground truth, before and after applying constrained decoding. The minimal differences suggest that constrained decoding does not compromise model exploration capability.}
\label{fig:ce_constrained_decoding}
\end{figure}

\section{Implementation Details of Passage Retrieval for KG Exploration}
\label{sec:implementation}

Following the implementation of \cite{sun2024thinkongraph}, our KG exploration framework adopts a lightweight retrieval module at each step to select relevant candidates from a predefined list. Given a question $q$, and a list of candidate passages $P_{cand}$ (either relation phrases or entity names), the goal of retrieval is to identify the top-$k$ most relevant candidates that guide the next reasoning step.

\subsection*{Retrieval Formulation}

For each step, we compute a relevance score between the question $q$ and every candidate passage $p \in P_{\text{cand}}$. The top-$k$ passages with the highest scores are selected:
\[
P_q = \text{Top}_k \left( \text{score}(p, q) \right), \quad \forall p \in P_{\text{cand}}.
\]
The scoring function $\text{score}(p, q)$ depends on the retrieval method used (BM25 or embedding-based retrievers).

\subsection*{BM25 Retriever}

For keyword-based retrieval, we use BM25 via the \texttt{rank\_bm25} implementation. Each passage (e.g., a relation like ``place of birth'' or an entity name like ``Albert Einstein'') is treated as a short bag-of-words document. The question $q$ is tokenized into a word list $q_1,\cdots,q_n$, and its relevance to each passage is computed based on term frequency and inverse document frequency:
\[
\text{score}(p, q) = \text{BM25}(p, q)
\]

\subsection*{Embedding-Based Retrievers}

For embedding-based retrievers such as SentenceBERT and GTR, we encode both the question and candidate passages using a pretrained text encoder $\mathcal{T}(\cdot)$. The relevance score is computed as the dot product between their embeddings:
\[
\text{score}(p, q) = \langle \mathcal{T}(p), \mathcal{T}(q) \rangle.
\]

\section{Constrained Decoding with JSON Format}
\label{sec:json}

To ensure that the performance gap between SLMs and LLMs is not simply due to formatting inconsistencies or output mismatches, we adopt a constrained decoding strategy across all models. Specifically, we modify the prompts to require all models to produce answers strictly in a predefined JSON format. Comparisons of original prompt and our modified prompt are showed in Table~\ref{tab:relation_prompt_comparison} and ~\ref{tab:entity_prompt_comparison}.

By enforcing the constrained output structure, we ensure that all models, regardless of size, are evaluated under consistent conditions. We also conducted a quantitative analysis of relation cleaning errors before and after applying constrained decoding. Specifically, we counted how many times the model-generated outputs contained unparseable relation entries. As shown in Figure~\ref{fig:relation_cleaning_error_comparison}, constrained decoding substantially reduces relation formatting errors, especially for smaller models like Qwen2-0.5b and Qwen2-1.5b. This confirms that our constrained format enforcement effectively standardizes model outputs and mitigates noisy relation representations, allowing us to more reliably evaluate reasoning quality.

After removing parsing-related noise, we further examined whether the adoption of constrained decoding negatively impacts the LLMs' exploration ability. To assess this, we computed the cross entropy (CE) between the retrieved relation paths and the ground-truth paths under both the original and constrained prompt settings. 

As shown in Figure~\ref{fig:ce_constrained_decoding}, the CE values remain stable across models, with negligible changes before and after applying constrained decoding. This result confirms that our constrained decoding strategy effectively removes parsing-related variance without diminishing the LLMs’ ability to explore and select relevant paths.

\begin{table}
  \centering
  \begin{tabular}{l|c|c}
    \hline
    Models & CWQ & WebQSP \\
    \hline
    GPT-4.1 & \textbf{0.575} & \textbf{0.810} \\
    $\quad$w/ BM25 & 0.525 & 0.745 \\
    $\quad$w/ SentenceBERT & 0.520 & 0.775 \\
    $\quad$w/ GTR & 0.505 & 0.805 \\ \hline
  \end{tabular}
    \caption{The performance of GPT-4.1 equipped with different exploration modules.}
\label{tab:llm_retrieval}
\end{table}

\begin{table*}[t]
\centering
\begin{tabular}{|p{0.95\textwidth}|}
\hline
\textbf{\small Original Extract Relation Prompt (Unconstrained)} \\
\hline
\begin{lstlisting}[basicstyle=\small\rmfamily,lineskip=0.1ex,aboveskip=0.2ex,belowskip=0.2ex,breaklines=true,columns=fullflexible ]
Please retrieve 3 relations (separated by semicolon) that contribute to the question and rate their contribution on a scale from 0 to 1 (the sum of the scores of %s relations is 1).

Q: Name the president of the country whose main spoken language was Brahui in 1980?
Topic Entity: Brahui Language
Relations: language.human_language.main_country; language.human_language.language_family; language.human_language.iso_639_3_code; base.rosetta.languoid.parent; language.human_language.writing_system; base.rosetta.languoid.languoid_class; language.human_language.countries_spoken_in; kg.object_profile.prominent_type; base.rosetta.languoid.document; base.ontologies.ontology_instance.equivalent_instances; base.rosetta.languoid.local_name; language.human_language.region
A: 
1. {language.human_language.main_country (Score: 0.4))}: This relation is highly relevant as it directly relates to the country whose president is being asked for, and the main country where Brahui language is spoken in 1980.
2. {language.human_language.countries_spoken_in (Score: 0.3)}: This relation is also relevant as it provides information on the countries where Brahui language is spoken, which could help narrow down the search for the president.
3. {base.rosetta.languoid.parent (Score: 0.2)}: This relation is less relevant but still provides some context on the language family to which Brahui belongs, which could be useful in understanding the linguistic and cultural background of the country in question.
Q:
\end{lstlisting}\\
\hline
\textbf{\small Modified Extract Relation Prompt (Constrained Decoding)} \\
\hline
\begin{lstlisting}[basicstyle=\small\rmfamily,lineskip=0.2ex,aboveskip=0ex,belowskip=0ex,breaklines=true,columns=fullflexible ]
Please retrieve 3 relations that contribute to the question and rate their contribution on a scale from 0 to 1 (the sum of the scores of 3 relations is 1). Provide the output in JSON format.

Q: Name the president of the country whose main spoken language was Brahui in 1980?
Topic Entity: Brahui Language
Relations: language.human_language.main_country; language.human_language.language_family; language.human_language.iso_639_3_code; base.rosetta.languoid.parent; language.human_language.writing_system; base.rosetta.languoid.languoid_class; language.human_language.countries_spoken_in; kg.object_profile.prominent_type; base.rosetta.languoid.document; base.ontologies.ontology_instance.equivalent_instances; base.rosetta.languoid.local_name; language.human_language.region
A: 
{
  "relations": [
    {
      "relation": "language.human_language.main_country",
      "score": 0.4,
      "description": "This relation is highly relevant as it directly relates to the country whose president is being asked for, and the main country where Brahui language is spoken in 1980."
    },
    {
      "relation": "language.human_language.countries_spoken_in",
      "score": 0.3,
      "description": "This relation is also relevant as it provides information on the countries where Brahui language is spoken, which could help narrow down the search for the president."
    },
    {
      "relation": "base.rosetta.languoid.parent",
      "score": 0.2,
      "description": "This relation is less relevant but still provides some context on the language family to which Brahui belongs, which could be useful in understanding the linguistic and cultural background of the country in question."
    }
  ]
}
Q: 
\end{lstlisting}
\\
\hline
\end{tabular}
\caption{Comparison of original prompt and our constrained decoding version for relation pruning. The modified prompt enforces a strict JSON structure to enable consistent and parseable outputs from SLMs.}
\label{tab:relation_prompt_comparison}
\end{table*}

\begin{table*}[t]
\centering
\begin{tabular}{|p{0.95\textwidth}|}
\hline
\textbf{\small Original Score Entity Candidates Prompt (Unconstrained)} \\
\hline
\begin{lstlisting}[basicstyle=\small\rmfamily,lineskip=0.1ex,aboveskip=0.2ex,belowskip=0.2ex,breaklines=true,columns=fullflexible ]
lease score the entities' contribution to the question on a scale from 0 to 1 (the sum of the scores of all entities is 1).

Q: The movie featured Miley Cyrus and was produced by Tobin Armbrust?
Relation: film.producer.film
Entites: The Resident; So Undercover; Let Me In; Begin Again; The Quiet Ones; A Walk Among the Tombstones
Score: 0.0, 1.0, 0.0, 0.0, 0.0, 0.0
The movie that matches the given criteria is "So Undercover" with Miley Cyrus and produced by Tobin Armbrust. Therefore, the score for "So Undercover" would be 1, and the scores for all other entities would be 0.

Q: {}
Relation: {}
Entites:
\end{lstlisting}\\
\hline
\textbf{\small Modified Score Entity Candidates Prompt (Constrained Decoding)} \\
\hline
\begin{lstlisting}[basicstyle=\small\rmfamily,lineskip=0.2ex,aboveskip=0ex,belowskip=0ex,breaklines=true,columns=fullflexible ]
Please score each entity's contribution to the question on a scale from 0 to 1 (the sum of the scores of all entities should be 1). Provide the output in JSON format.

Q: The movie featured Miley Cyrus and was produced by Tobin Armbrust?
Relation: film.producer.film
Entities: The Resident; So Undercover; Let Me In; Begin Again; The Quiet Ones; A Walk Among the Tombstones

A: {{
  "entities": [
    {{"name": "The Resident", "score": 0.0}},
    {{"name": "So Undercover", "score": 1.0}},
    {{"name": "Let Me In", "score": 0.0}},
    {{"name": "Begin Again", "score": 0.0}},
    {{"name": "The Quiet Ones", "score": 0.0}},
    {{"name": "A Walk Among the Tombstones", "score": 0.0}}
  ],
  "explanation": "The movie that matches the given criteria is \"So Undercover,\" which features Miley Cyrus and was produced by Tobin Armbrust. Therefore, the score for \"So Undercover\" is 1, and the scores for all other entities are 0."
}}

Q: {}
Relation: {}
Entities:

\end{lstlisting}
\\
\hline
\end{tabular}
\caption{Comparison of original prompt and our constrained decoding version for entities pruning. The modified prompt enforces a strict JSON structure to enable consistent and parseable outputs from SLMs.}
\label{tab:entity_prompt_comparison}
\end{table*}

\section{Passage Retrieval for LLMs}
\label{sec:llm_retrieval}
In an ablation study conducted by \citet{sun2024thinkongraph}, they showed that using lightweight passage retrieval models for exploration significantly reduced the number of LLM calls from $2ND+D+1$ to $D+1$ where $D$, $N$ are the numbers of iterations and reasoning paths respectively. However, this efficiency gain came at the cost of a substantial drop in EM score. We reproduce the results in Table~\ref{tab:llm_retrieval}. In contrast, our experiments in Section~\ref{sec:slm_retrieval} demonstrate that passage retrieval models can offer the best of both worlds for SLMs: not only do they improve the efficiency of ToG, but they also enhance the EM performance, without facing the trade-off observed in the original study. The main reason for this difference in findings lies in the disparity between LLMs and SLMs in their ability to perform KG exploration. Therefore, their results complement, rather than contradict our findings.

\end{document}